\documentclass[12pt]{article}
\usepackage[margin=1in]{geometry}
\usepackage[utf8]{inputenc}
\usepackage{comment}
\usepackage{amsmath}
\usepackage{url}
\usepackage{amssymb}
\usepackage{amsthm, thmtools, thm-restate}
\usepackage{amsfonts}
\usepackage{bbm}
\usepackage{graphicx}
\usepackage[most]{tcolorbox}

\declaretheorem[name=Definition]{definition}

\usepackage{hyperref}
\hypersetup{
    colorlinks=true,
    linkcolor=blue,
    filecolor=red,      
    urlcolor=magenta,
    pdfpagemode=FullScreen,
    }

\usepackage[doi=false, url=false, isbn=false, eprint=false, backend=biber,style=authoryear,maxcitenames=3, maxbibnames=99]{biblatex}
\title{Operationalising Representation in Natural Language Processing}
\author{Jacqueline Harding}
\date{Stanford University\footnote{Correspondence to: hardingj@stanford.edu}}

\begin{document}

\maketitle

\section*{Abstract}

Neural models achieve high performance on a variety of natural language processing (NLP) benchmark tasks. How models perform these tasks, though, is notoriously poorly understood. As models are deployed more widely in real-world settings, there has been increased focus on \textit{explaining} model behaviour. Since the task the model performs is linguistic, it seems intuitive that models perform tasks by \textit{representing} linguistic properties of inputs.

Despite its centrality in the philosophy of cognitive science, there has been little prior philosophical work engaging with `representation' in contemporary NLP practice. This paper attempts to fill that lacuna: drawing on ideas from the philosophy of cognitive science, I introduce a framework for evaluating the representational claims made about components of neural NLP models, proposing three criteria with which to evaluate whether a component of a model represents a property and operationalising these criteria using `probing classifiers', a popular analysis technique in NLP (and deep learning more broadly).

The project of operationalising a philosophically-informed notion of representation should be of interest to both philosophers of science and NLP practitioners. It provides philosophers insight into the connections between empirical model interpretability work and foundational questions in cognitive science, assisting them in engaging with downstream philosophical issues in the development of AI. It helps NLPers organise the large literature on probing experiments, suggesting novel avenues for empirical research.

\tableofcontents

\section{Introduction}

Neural models\footnote{At times, I will restrict my discussion to large pre-trained transformer models (\cite{Vaswani2017}), such as BERT (\cite{Devlin2018}), RoBERTa (\cite{Liu2019}), GPT (\cite{Radford2018}, \cite{Radford2019}, \cite{Brown2020}), T5 (\cite{Raffel2020}) and BART (\cite{Lewis2020}). That said, I intend my arguments to apply to a wide class of neural models, so I aim to stay neutral as to the precise model under discussion.} achieve high performance on a variety of natural language processing (NLP) benchmark tasks (\cite{Wang2019}). How models perform these tasks, though, is notoriously poorly understood. As models are deployed more widely in real-world settings, there has been increased focus on \textit{explaining} model behaviour. Since the task the model performs is linguistic, it seems intuitive that models perform tasks by \textit{representing} linguistic properties of inputs.

Despite its centrality in the philosophy of cognitive science, there has been little prior philosophical work engaging with `representation' in contemporary NLP practice. This paper attempts to fill that lacuna: drawing on ideas from the philosophy of cognitive science, I introduce a framework for evaluating the representational claims made about components of neural NLP models, proposing three criteria with which to evaluate whether a component of a model represents a property and operationalising these criteria using `probing classifiers' (\cite{Alain_Bengio2016}), a popular analysis technique in NLP (and deep learning more broadly).

The project of operationalising a philosophically-informed notion of representation should be of interest to both philosophers of science and NLP practitioners. It provides philosophers insight into the connections between empirical model interpretability work and foundational questions in cognitive science, assisting them in engaging with downstream philosophical issues in the development of AI. It helps NLPers organise the large literature on probing experiments, suggesting novel avenues for empirical research.

The paper is organised as follows. In the remainder of this section, I motivate the need for a notion of representation suitable for NLP (Section \ref{motivation: two examples}) and introduce terminology for describing the setting of interest (Section \ref{general setting}). In Section \ref{representation in NLP}, I give (Section \ref{three criteria}) three plausible criteria for a component of a system to count as representing a property of an input and introduce (Section \ref{probing classifiers}) probing classifiers, a popular technique for analysing neural models which serves as a basis for an operationalisation of the criteria. (Much of) Section \ref{three criteria} will be familiar to philosophers, whilst Section \ref{probing classifiers} will be familiar to NLPers. Section \ref{operationalising the three criteria} contains the main contribution of the paper, in which I operationalise the three criteria (\hyperlink{information criterion}{Information}, \hyperlink{use criterion}{Use} and \hyperlink{misrepresentation criterion}{Misrepresentation}) via definition of two sorts of interventions on model activations. In Section \ref{discussion}, I draw connections between my operationalisation of representation and the empirical NLP literature; this section will be of particular interest to NLPers, but it also serves as a relatively self-contained introduction to several interpretability techniques which have received recent interest from philosophers. I conclude (Section \ref{conclusion}) by summarising the contributions of the paper, arguing that they should interest both philosophers and NLPers.

\subsection{Motivation: Two Examples}\label{motivation: two examples}

To demonstrate the need for an account of representation in NLP, I outline the following toy examples.

\subsubsection{Example 1}\hypertarget{example 1}{}

Suppose BERT (\cite{Devlin2018}) is fine-tuned on the Corpus of Linguistic Acceptability (CoLA; \cite{warstadt2018}). CoLA is a binary classification task, where the model judges whether an input sentence is grammatical or not.

BERT has been pre-trained on a large document-level corpus on two self-supervised tasks: a masked language modelling task and a next sentence prediction task. For the fine-tuning stage, a linear classifier is appended to the end of the model. The classifier takes in the activation above the first token from BERT's final layer and predicts a probability distribution over labels $C = \{\texttt{grammatical}, \texttt{ungrammatical}\}$. The combination of model and classifier are tuned (a supervised task) on the labelled training portion of CoLA.

Suppose that at test time BERT incorrectly judges the following CoLA example as ungrammatical:
\begin{quote}
    \texttt{John danced waltzes across the room.}
\end{quote}
A researcher suggests the following high-level explanation for BERT's error: the model judges  `waltzes' to be a verb, when it is in fact a noun.

\subsubsection{Example 2}\hypertarget{example 2}{}

GPT-3 (\cite{Brown2020}) is performing a zero-shot co-reference resolution task, such as the Winograd Schema Challenge (\cite{Levesque2012}).

Like BERT, GPT-3 has been pre-trained on a large document-level corpus. Unlike BERT, GPT has been trained on a standard left-to-right language modelling task (using \citeauthor{Vaswani2017}'s terminology, BERT is constructed from stacked `encoder' transformer blocks whereas GPT is constructed from stacked `decoder' blocks). Since GPT-3 is performing the task using in-context learning, there is no additional component to the system (but rather an `unembedding' layer which outputs a distribution over the vocabulary), and no fine-tuning occurs.

We notice (similarly to \cite{Brown2020}, who test GPT-3 on a gendered version of Winograd from \cite{Rudinger2018}) that for the (fictional) input
\begin{quote}
    \texttt{The doctor told the patient \textit{she} was unwell. Who was unwell?}
\end{quote}
GPT-3 suggests the completion `the patient', whereas for the input
\begin{quote}
    \texttt{The doctor told the patient \textit{he} was unwell. Who was unwell?}
\end{quote}
GPT-3 suggests the completion `the doctor'.

We hypothesise that this is because GPT-3 \textit{represents} the doctor as male (and -- perhaps -- the patient as female).

\subsubsection{Discussion of Examples 1 and 2}

Both examples contain partial explanations of model behaviour. Because these explanations involve the model responding to (and manipulating) properties of inputs to perform complex linguistic tasks -- tasks for which there do not exist perspicuous non-representational explanations (\cite{vanGelder1995}) -- they are naturally seen as making claims about what the model \textit{represents}. These are instances of a more general phenomenon: explanations of contemporary artificial neural models which purport to be human-understandable often presuppose a notion of a model representing a property (\cite{buckner2019}, \cite{Cao2022}, \cite{elhage2021mathematical}, \cite{Lipton2016}).

To evaluate the claim made in \hyperlink{example 1}{Example 1}, we need an account of what it means for BERT to (mis)judge a word to be a noun or verb: that is, we need to understand what it means for BERT to \textit{represent} words' part of speech (PoS) across some range of inputs, and what it means for BERT to \textit{misrepresent} the word `waltzes' as a verb in this particular case. In \hyperlink{example 2}{Example 2}, we need to know what it is for a model to represent doctors as male, or patients as female; downstream work which attempts to `de-bias' a model's internal states (\cite{Bolukbasi2016}) implicitly makes a similar representational claim.

It is important to make three clarifications at this stage. First, throughout this paper, I'm using `representation' in the sense that philosophers of cognitive science do, to denote a concept intimately connected with explanation (it is this sense of representation that supports counterfactuals like `if the system $S$ hadn't represented property $Z$, it would have failed to perform the task'). I'm concerned only with subpersonal representations (representations of properties of inputs by components of systems), not personal-level representations. Note also that there is a thinner use of the word `representation' -- ubiquitous in machine learning -- in which any internal state of a neural model counts as a `representation' (i.e. `representation' is often used as a synonym for `intermediate activation'). My point isn't merely sociological (that philosophers and NLPers use the word `representation' differently), but rather that many explanations in NLP presuppose the sense of representation used by philosophers of cognitive science.

Second, the candidate explanations need not actually be true (or even plausible) for the claim I'm making to hold. Indeed, even in the toy example tasks, any actual explanation of model behaviour will be much more complex than the explanations above. The point is rather that evaluating the explanations' truth requires an account of what it means for a model to represent a property (so all that is needed is plausible truth-\textit{aptness}). Moreover, this account should be \textit{operationalisable}: given the account, it should (in principle) be possible to test for a representation of a property.

Third, nothing said here commits me to the claim that models actually have internal representations of linguistic properties (although in the context of contemporary models I find this claim compelling). It might be that any sense we have of models `representing' language based on their behaviour is entirely misleading. This would constitute an \textit{error theory} about representational explanations. To put it in Marr's (\citeyear{marr1982vision}) terms, the thought would be that -- to the extent that we could explain models' behaviour -- such explanations would sit only at the implementational level (in terms of the way models manipulate points in a high-dimensional space) or at the computational level (in terms of input/output mappings), but not at the algorithmic level. Again, though, entering into this discussion presupposes an account of representation.

\subsection{The General Setting}\label{general setting}

\begin{figure}
    \centering
    \includegraphics[scale=0.3]{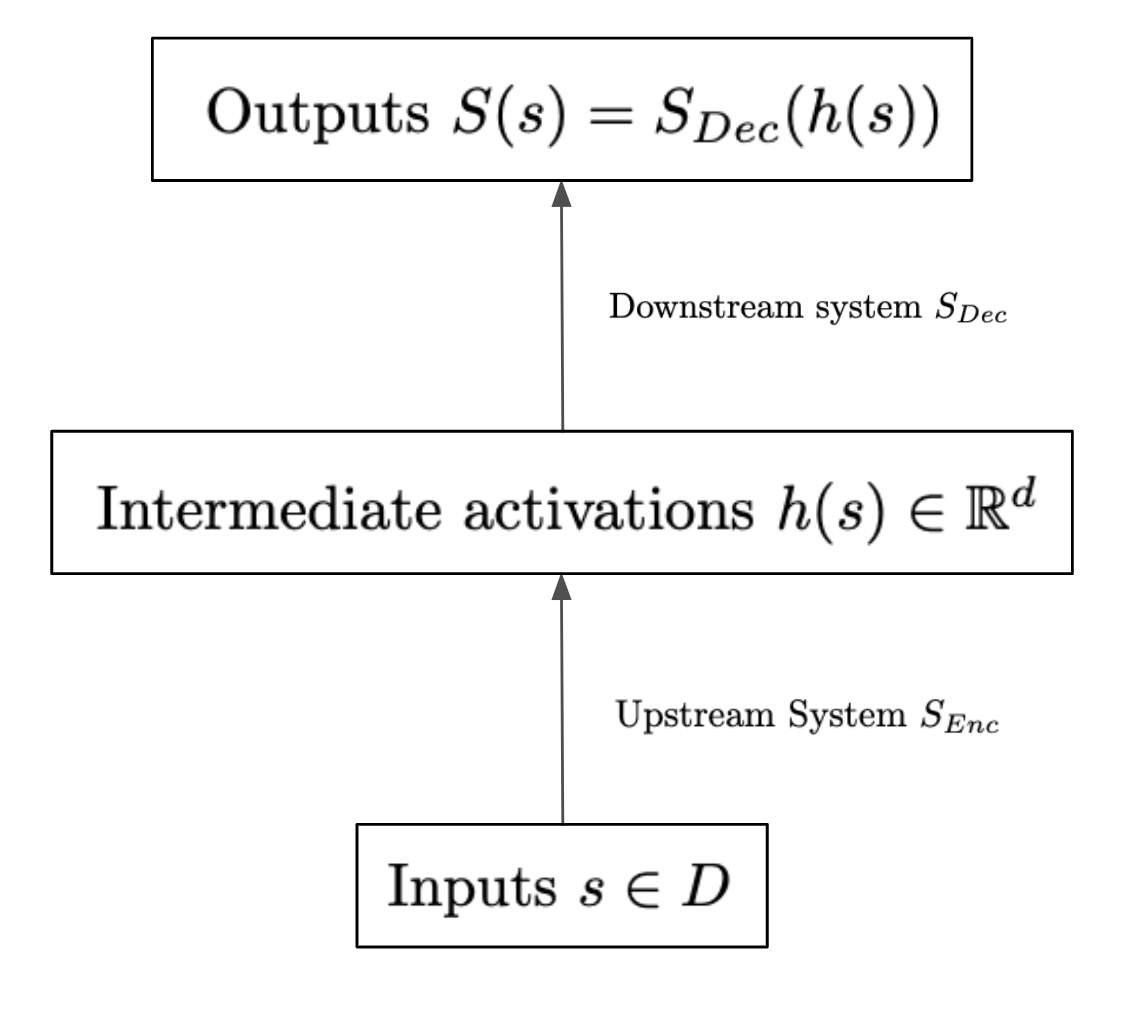}
    \caption{The General Setting: a System Performing a Task}
    \label{fig:general_setting}
\end{figure}

In the general setting (depicted in Figure \ref{fig:general_setting}), we have a system $S$ performing a task $D$. The system $S$ takes an input $s$ and maps it to an output $S(s)$. For simplicity, I treat the task $D$ as a finite set of inputs $s \in D$ along with an associated `goodness measure' on outputs $S(s)$ (for example, a loss function); so different outputs of the system are `better' or `worse' relative to the task $D$.

In the simplest case, we can see the system $S$ as the composition of two functions: an `upstream system' which takes an input $s$ and maps it to an `intermediate activation vector' $h(s)$ (which I assume is a $d$-dimensional real vector), and a `downstream system' which takes the intermediate activation vector $h(s)$ and maps it to an output $S(s)$. To foreshadow the role that this decomposition will play in my account of representation, I call the upstream system $S_{Enc}$ (for `encoder') and the downstream system $S_{Dec}$ (for `decoder').

The setting here is general enough to accommodate a variety of neural models and tasks. More concretely, we could think of $S$ as an autoregressive transformer (\cite{Vaswani2017}), inputs $s$ as finite lists of subword \textit{tokens} from some fixed vocabulary, each $h(s)$ as the value of the \textit{residual stream} (\cite{elhage2021mathematical}) at some layer and position in the context window, $S_{Enc}$ as the layers below $h$, $S_{Dec}$ as the layers above $h$, and outputs $S(s)$ as probability distributions over next tokens.

Let $Z$ be a finite set of `property labels'. More concretely, $Z$ could be a property of individual words (e.g. a set of PoS tags -- as in \hyperlink{example 1}{Example 1} -- or dependency heads) or of whole sentences (a binary variable corresponding to whether the subject of the sentence is singular or plural). Although many of the properties about which representational claims are made in NLP are syntactic, $Z$ could also be a more abstract property (e.g. a set of labels for the gender of the subject of the sentence, as in \hyperlink{example 2}{Example 2}, or a set of sentiment labels for the sentence). I assume each input $s$ is associated with a `true label distribution' $p_{Z}(s)$ over the labels in $Z$ (that is, $p_{Z}(s) \in \mathcal{P}(Z)$, where $\mathcal{P}(Z)$ denotes the set of probability mass functions over $Z$).

When the system $S$ processes inputs $s \in D$, we get a \textit{pattern of activations} $h(D)$, defined as follows:
\begin{equation*}
    h(D) = \{h(s) \:|\: s \in D\}
\end{equation*}
(i.e. a pattern of activations is just a set of intermediate activation vectors, one for each input comprising the task).

Contemporary accounts of representation distinguish between a representational \textit{vehicle} and the information it bears about the input $s$ (its representational \textit{content}). The content supervenes on the vehicle that bears it; it is the vehicle which interacts with the rest of the system, and it is via this interaction that information passes through the system. In neural models, a natural vehicle for content is the pattern of activations $h(D)$.

Having motivated the need for an account of representation in NLP and specified the general setting, I turn to the question this paper seeks to answer: what does it mean for the pattern of activations $h(D)$ to represent property $Z$ of inputs $s$?

\section{Representation in NLP}\label{representation in NLP}

In this section, I give (Section \ref{three criteria}) three plausible criteria for a component of a system to count as representing a property of an input. I then introduce (Section \ref{probing classifiers}) probing classifiers, a popular technique for analysing neural models which I suggest can help operationalise the criteria.

\subsection{Three Criteria}\label{three criteria}

I suggest that a pattern of activations $h(D)$ \textit{represents} a property $Z$ if the following three criteria hold:

\begin{restatable}[Information]{definition}{information}\hypertarget{information criterion}
The pattern of activations $h(D)$ bears information about $Z$.
\end{restatable}

\begin{restatable}[Use]{definition}{use}\hypertarget{use criterion}
The information the pattern of activations $h(D)$ bears about $Z$ is used by system $S$ to perform task $D$.
\end{restatable}

\begin{restatable}[Misrepresentation]{definition}{misrepresentation}\hypertarget{misrepresentation criterion}
It should be possible (at least in principle) for the activation vector $h(s)$ to \textit{misrepresent} $Z$ with respect to an input $s \in D$.
\end{restatable}

\noindent By specifying these criteria, I don't take myself to be proposing a new theory of representation in cognitive science, nor am I evaluating existing accounts of representation by how amenable they are to operationalisation. My goal in this paper is rather to take conditions on representation widely endorsed by philosophers of cognitive science and operationalise them in an NLP setting.\footnote{In this, I follow projects applying philosophically-informed notions of representation to analysis techniques in neuroscience, such as \textcite{Roskies2021}.} One might label this an `operationalisation first' approach to representation.\footnote{This could be thought of as a species of `representational pragmatism' (\cite{Cao2022}).} My hope is that by resisting tying my operationalisation to a particular account of representation, proponents of different accounts can re-purpose my framework with their own theoretical paraphernalia.\footnote{I intend the operationalisation I introduce to be amenable to a wide variety of teleosemantic accounts of representation, including those which emphasise the role of a `producer' of the representation (\cite{Neander2013}, \cite{Neander2017}) and those which emphasise the role of a `consumer' of the representation (\cite{Millikan1984}, \cite{Millikan2004}), as well as `conceptual role' accounts of representation (\cite{Harman1982}, \cite{Block1987}, \cite{piantadosi2022meaning}).} Given that different accounts of representation will re-purpose the criteria differently, the criteria ought not to be thought of as necessary and sufficient conditions on representation, but rather important building blocks in a theory of representation.\footnote{This disclaimer is largely intended for the philosophers reading this paper. Given the centrality of (something like) these criteria in contemporary philosophical discussion of representation, NLPers interested in applying ideas from the philosophy of representation to their own work could in practice treat these criteria as crudely necessary and sufficient for representation in a wide range of experimental settings (with the caveat that there is there is nuanced, ongoing philosophical disagreement about the details involved in fleshing out the criteria).}

Although I take it that \hyperlink{information criterion}{Information}, \hyperlink{use criterion}{Use} and \hyperlink{misrepresentation criterion}{Misrepresentation} have broad philosophical appeal, I want briefly to justify their importance in assessing representational claims. It's helpful to give an example of an account of representation in which some version of the criteria play a role, namely `varitel semantics' (\cite{Shea2018-SHERIC}). Amongst other things, varitel semantics counts components of systems as representing properties when they carry `correlational information' which plays a `role in explaining' the system's performance of a `task function' (p.84). A `task function' is a sufficiently robust behaviour of the system which has been `stabilised' (p.64) by being selected for or learned.

For our purposes, we can see that the task $D$ (construed as a set of inputs along with a goodness measure on outputs) can be thought of as a task function, the system's performance of which we are seeking to explain. It's clear that \hyperlink{information criterion}{Information} requires that a component bears `correlational information'. Similarly, demonstrating \hyperlink{use criterion}{Use} holds is a first step in explaining a system's successful performance of the task.

Showing that the system uses the information a pattern of activations bears about a property to perform the task helps explain the system's \textit{successes} on the task. However, we are also interested in explaining the system's \textit{failures} on the task. This is especially important in the present context: a key motivation for explaining neural NLP models is to explain their failures (presumably with an eye to correcting them), as gestured at in Examples \hyperlink{Example 1}{1} and \hyperlink{Example 2}{2}. It is this idea which \hyperlink{misrepresentation criterion}{Misrepresentation} is intended to capture. If we explain a neural model's behaviour in representational terms, then one source of failure will be \textit{misrepresentation}, a case in which a particular activation vector $h(s)$ is an `incorrect representation' (of some particular property) relative to the system's task function (\cite{Shea2018-SHERIC}). Broadly speaking, a representation of a property is incorrect if, when interacted with in the usual way, it leads the downstream system to process the input as though the input has (on my terminology) a property label it does not actually have; such processing will typically -- though not always -- result in a degradation of the system's performance on the task, and it is the `incorrectness' of this behaviour which grounds the incorrectness of the representation. Incorrect representations can occur in a variety of ways; for example, (using my notation) something may go \textit{wrong} with the way the upstream system $S_{Enc}$ produces a particular activation vector $h(s)$, or it might be that no malfunction with the upstream system occurs but rather there has been some shift in the distribution of inputs (a case in which the environment is `uncooperative').\footnote{Of course, different accounts of representation will disagree as to which of these cases are misrepresentations. For example, proponents of `low-church' teleosemantics (\cite{Neander1995}) will usually deny that cases of uncooperative environments count as misrepresentations.}

Of course, the comments above are far from a full development of varitel semantics in the context of modern NLP; there are many interesting questions to explore.\footnote{To name two: I haven't discussed what it means for components of systems to `implement algorithms' or what it means for information to play an `unmediated role' in explaining the system's performance of tasks (\cite{Shea2018-SHERIC}, p.84). I intend these lines of development to be compatible with my project here.} For example, it's natural to observe that task functions are easier to spell out in contemporary machine learning models than biological ones (\cite{Cao2022}), in that artificial systems are trained by optimising an objective function using gradient descent; this observation motivates my stipulation of the task $D$. This appears to imply that the stabilisation process relevant for ascribing task functions to NLP systems is learning, but the current NLP training pipeline complicates this. As exemplified in Examples \hyperlink{Example 1}{1} and \hyperlink{Example 2}{2}, contemporary models go through both pre-training and tuning stages using different objective functions and optimisers, and it's unclear whether in-context learning (\cite{Brown2020}) ought to count as a stabilising process: the tasks $D$ whose performance we want to explain need not be tasks for which models were explicitly optimised. It's also worth observing that a selection process operates on populations of models; less successful models quickly fade into obsolescence.

In the context of modern NLP, the characterisation of the task function $D$ also bears on a foundational question: do `the internal representations and outputs of Large Language Models possess intrinsic meaning, enabling them to represent things in the world independently of human interpretation' (\cite{mollo2023vector}, p.8)? This is an updated version of the familiar `Symbol Grounding Problem' (\cite{Harnad_1990}), which -- given contemporary models' significantly improved performance on natural language tasks -- has received renewed attention (\cite{bender-koller-2020-climbing}, \cite{Pavlick2023}, \cite{mollo2023vector}, \cite{mandelkern2023language}).\footnote{Thanks to an anonymous referee for discussion on this point.}

Both \textcite{piantadosi2022meaning} and \textcite{Pavlick2023} suggest that conceptual role semantics (CRS; \cite{Harman1982}, \cite{Block1987}), whereby the content of representations is determined by their functional roles within the system, provides a natural framework for grounding meaning in contemporary neural NLP models. Of course, the plausibility of this suggestion as a complete solution to the grounding problem will depend on the sorts of properties we want to feature in our representational explanations. If we want to move beyond properties of inputs (e.g. linguistic properties) to properties of the world proper, then CRS must either allow for a referential component to meaning (a `two-factor' account) or make use of `long-armed' conceptual roles, those which stretch beyond the boundaries of the model;\footnote{See \cite{Block1986-BLOAFA-2} for discussion; in particular, \citeauthor{Block1986-BLOAFA-2} argues that long-armed one-factor accounts -- like \citeauthor{Harman1982}'s -- collapse into a two-factor account.} \citeauthor{Pavlick2023} (p.10) adopts the former approach, appealing to a causal theory to determine reference. Similarly, \textcite{mollo2023vector} assume that CRS cannot by itself resolve the referential component of the grounding problem. They suggest that in order to ground non-linguistic representational content, the task $D$ the model performs must itself involve the world; in particular, they suggest that contemporary tuning techniques encourage models to perform these `world-involving' tasks.\footnote{A version of this point -- though aimed at grounding the outputs of LLMs -- is also made by \textcite{mandelkern2023language}, who argue that the tasks LLMs perform ground their reference, via establishing them as members of the right sort of linguistic community.}

By stipulating the task $D$ (rather than searching for a world-involving task that is independent of human interpretation), my intention is to circumvent further engagement with the grounding problem; in effect, the notion of representation I operationalise is task-relative. This is because I see the motivation for my project as largely independent of the grounding problem. To see this, note that an operationalisation of representation is needed even if the grounding problem is solved: knowing that models \textit{can} represent the world successfully is not sufficient to determine \textit{what} features of the world they represent (indeed, the properties I discuss in this paper can all be construed in linguistic terms, which -- as discussed above -- would simplify the grounding problem). Furthermore, even if there were no interpretation-independent task which could ground meaning in contemporary NLP models (i.e. even if there were no solution to the grounding problem), it would still be necessary to have an account of representation which accounted for the utility of representational explanations; in such a case, we would still need a principled framework for understanding how to identify and manipulate representations (\cite{Cao2022}).

Future work should pay more attention to the subtleties here. The discussion above is intended to justify the importance of the three criteria I've introduced, so as to motivate the project of operationalising them. I turn now to an important tool in the operationalisation I propose.

\subsection{Probing Classifiers}\label{probing classifiers}

In this subsection, I introduce probing classifiers (\cite{ettinger-etal-2016-probing}, \cite{gupta-etal-2015-distributional}, \cite{kohn-2015-whats}, \cite{Alain_Bengio2016}), a popular analysis technique in deep learning.
\begin{figure}[h!]
    \centering
    \includegraphics[scale=0.3]{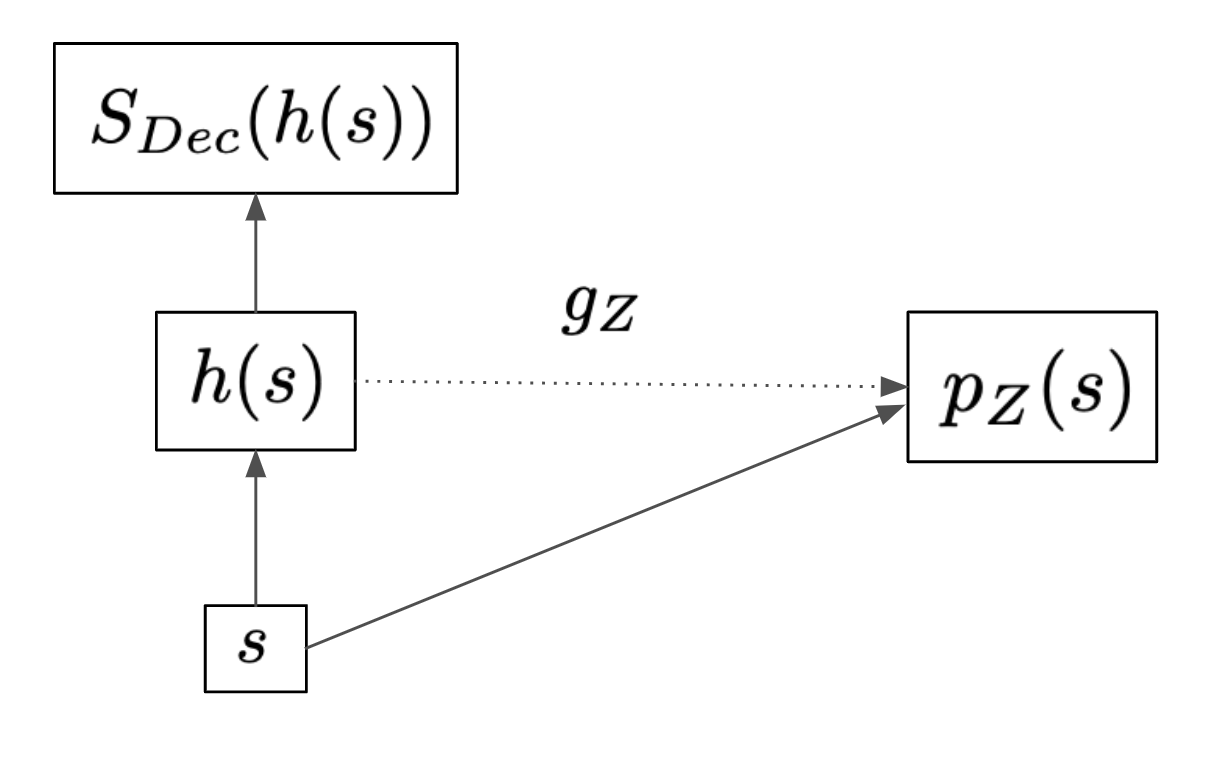}
    \caption{Training a Probe $g_{Z}$ for Property $Z$}
    \label{fig:probe_setting}
\end{figure}
Given a system $S$ performing a task $D$ and a component $h$ of $S$, a \textit{probe} for property $Z$ is a function $g_{Z}:h(D)\rightarrow \mathcal{P}(Z)$; that is, a classifier which takes in (for inputs $s \in D$) an intermediate activation vector $h(s)$ and outputs a label distribution $g_{Z}(h(s))\in \mathcal{P}(Z)$. This situation is depicted in Figure \ref{fig:probe_setting}.

Probes are trained to predict the true label distribution $p_{Z}(s)$ in a supervised way. We train $g_{Z}$ on a labelled dataset consisting of pairs
\begin{equation*}
    \{(h(s), p_{Z}(s))\}_{s \in D_{g_{train}}}
\end{equation*}
for a training set $D_{g_{train}}\subseteq D$, by minimising the cross-entropy loss (\cite{Cover_Thomas_2006})
\begin{equation*}
    L_{CE}(p_{Z}(s), g_{Z}(h(s)))
\end{equation*}
between the true label distribution $p_{Z}(s)$ and the probe's predictions $g_{Z}(h(s))$. The probe $g_{Z}$ is tested on a test set consisting of pairs
\begin{equation*}
    \{(h(s), p_{Z}(s))\}_{s \in D_{g_{test}}}
\end{equation*}
for inputs $D_{g_{test}}\subseteq D$ disjoint from $D_{g_{train}}$.
\begin{definition}[Successful Probe]\label{successful probe}
    A probe $g_{Z}$ for property $Z$ is \textit{successful} when its average loss on the test set
    \begin{equation*}
        \frac{1}{|D_{g_{test}}|}\sum_{s \in D_{g_{test}}} L_{CE}(p_{Z}(s), g_{Z}(h(s)))
    \end{equation*}
    is below some threshold.
\end{definition}
\noindent There is a large literature applying probing in NLP (\cite{shi-etal-2016-string}, \cite{Adi}, \cite{belinkov-etal-2017-evaluating_layers_of_NMT}, \cite{conneau-etal-2018-cram}, \cite{zhang-bowman-2018}, \cite{Hupkes2018VisualisationA}, \cite{liu-etal-2019-linguistic}, \cite{tenney2018what}, \cite{saleh-etal-2020-probing}; see \cite{belinkov2021probing} for a survey). There is also ongoing discussion about how best to baseline probing experiments and select among successful probes (\cite{zhang-bowman-2018}, \cite{liu-etal-2019-linguistic}, \cite{Pimentel2020}, \cite{Hewitt-Liang-2019}, \cite{pimentel-etal-2020-pareto}, \cite{voita-titov-2020-information}, \cite{pimentel-cotterell-2021-bayesian}). In Section \ref{discussion}, I argue that the framework I introduce helps make sense of this literature.

\section{Operationalising the Three Criteria}\label{operationalising the three criteria}

In this section, I will operationalise the three criteria introduced in Section \ref{representation in NLP}.

\subsection{Operationalising the Information Criterion}\label{operationalising information criterion}

Recall that \hyperlink{information criterion}{Information} was defined as follows:
\information*

\noindent \hyperlink{information criterion}{Information} is straightforward to operationalise using probing classifiers.

Let \textbf{s} be a random variable over inputs, with a distribution $P_{\textbf{s}}$ induced by the task $D$.\footnote{By `induced by the task $D$', I intend to capture the intuitive idea that the distribution of inputs depends on the setting in which we evaluate the model (for example, the inputs we would expect to see in a sentiment analysis setting are different from the inputs in a code completion setting).} Let $\mathbf{Z}$ be a random variable over property labels $Z$, with a distribution $P_{\mathbf{Z}}$ induced by the task $D$ (it might help to think of the gold label distribution $p_{Z}(s)$ as the result of conditionalising $P_{\mathbf{Z}}$ on the event $\mathbf{s}=s$). Similarly, let $\mathbf{h}$ be a random variable over activation vectors $h(s)$, with distribution $P_{\mathbf{h}}$ induced by $P_{\textbf{s}}$.

A natural first step towards operationalising \hyperlink{information criterion}{Information} is to say that a pattern of activations $h(D)$ bears information about a property $Z$ just when the Shannon mutual information $I(\textbf{Z};\textbf{h})$ (\cite{Cover_Thomas_2006}) between the random variables $\textbf{Z}$ and $\textbf{h}$ is sufficiently high (intuitively, if learning the value of the activation vector $h(s)$ sufficiently reduces uncertainty about the label distribution of input $s$, given the pattern of activations $h(D)$).\footnote{See \textcite{Usher2001} for an partial operationalisation of representation using mutual information. It varies significantly from my account in that it is not developed in the context of neural models, and does not involve causal interventions.}

Given this step, we can see (following \textcite{Pimentel2020}) that the cross-entropy loss of a probe $g_{Z}$ for property $Z$ lower bounds precisely the quantity we're interested in. First, write the mutual information between $\textbf{Z}$ and $\textbf{h}$ as
\begin{equation*}
    I(\textbf{Z};\textbf{h}) = H(\textbf{Z}) - H(\textbf{Z} \:|\: \textbf{h})
\end{equation*}
where $H(-)$, $H(-\:|\:-)$ denote entropy and conditional entropy, respectively (\cite{Cover_Thomas_2006}). Then it is easy to see that the (expected) empirical cross-entropy loss $L_{CE}$ of probe $g_{Z}$ upper bounds the conditional entropy $H(\textbf{Z} \:|\: \textbf{h})$ (see \textcite{zhu-rudzicz-2020} for a proof of this). So finding a \textit{successful} probe $g_{Z}$ -- a probe whose expected loss is below some threshold (Definition \ref{successful probe}) -- guarantees that $I(\textbf{Z};\textbf{h})$ is above some threshold (where this second threshold will depend on the entropy $H(\textbf{Z})$). This suggests an operationalisation of \hyperlink{information criterion}{Information} as follows:
\begin{restatable}[Information Operationalised]{definition}{information-operationalised}\hypertarget{information-o}{}
    $h(D)$ bears information about $Z$ iff we can find a successful probe $g_{Z}:h(D) \rightarrow \mathcal{P}(Z)$.
\end{restatable}

\subsection{Operationalising the Use Criterion}\label{operationalising use criterion}

In this subsection, I operationalise \hyperlink{use criterion}{Use}. I suggest that we can test whether the system uses information about a property by performing an appropriate intervention on the system's intermediate activations and assessing its effects on the system's outputs.\footnote{A large body of work in NLP takes a similar causal perspective. See (e.g.) \cite{Elazar2020}, \cite{feder-etal-2021-causalm}, \cite{Giulianelli2018}, \cite{Tucker2021}, \cite{Tucker2022}, \cite{vig2020_causalmediation}, \cite{geiger2019}; I discuss extant work in Section \ref{causal interventions}.}

Recall that \hyperlink{use criterion}{Use} was defined as follows:
\use*
\noindent Suppose we have a pattern of activations $h(D)$ which bears information about a property $Z$. Given a particular input $s$ and activation vector $h(s)$, the question is: what does it mean for the system to count as using the information $h(s)$ bears about $Z$ to perform task $D$?

There's a problem with this question as posed: although it's natural to speak of individual activation vectors $h(s)$ bearing information about $Z$, this way of speaking is misleading. In particular, it doesn't make sense to ask about the mutual information a single activation vector $h(s)$ bears about a property $Z$; this is akin to asking about the trivial quantity $I(\mathbf{Z};\mathbf{h'})$, where $\mathbf{h'}$ is a constant random variable (which takes value $h(s)$ with probability $1$).

This is an instance of a more general principle: individual messages are only informative relative to a decoding scheme. To the extent that it makes sense to speak of a single activation vector $h(s)$ bearing information, then, it is in virtue of the fact that $h(s)$ is part of a \textit{pattern of activations} $h(D)$; it is $h(D)$ that is the primary vehicle for informational content. So the information $h(s)$ bears about $Z$ can only be defined relative to a \textit{way of decoding} the information that $h(D)$ bears about $Z$.\footnote{A related notion is that of `usable information', in which mutual information is relativised to a `predictive family' of decoders. This is introduced by \textcite{Xu2020-usable-information} (and applied in the context of probing by \cite{Hewitt-ConditionalProbing-2021}). Their framework applies only to information carried by patterns of activations $h(D)$, rather than individual activation vectors.} A way of decoding the information $h(D)$ bears about $Z$ is just a successful probe $g_{Z}:h(D)\rightarrow \mathcal{P}(Z)$.

Fixing some successful probe $g_{Z}$, the discussion above motivates the following definition:
\begin{definition}[Activation Vector Bearing Information Relative to a Probe]\label{information relative to a probe}
    The amount of information an activation vector $h(s)$ bears about $Z$ relative to a probe $g_{Z}$ is given by the negative entropy
    \begin{equation*}
        -H(g_{Z}(h(s)))
    \end{equation*}
    of the probe's predictions on the activation vector.
\end{definition}
\noindent This definition captures the intuitive idea that the more `certain' the probe's predictions are, the more information about $Z$ the activation vector $h(s)$ provides relative to the probe, and vice-versa. When $g_{Z}(h(s))$ is uniform (so $-H(g_{Z}(h(s))) = - \log_{2}(|Z|)$ is minimal), $h(s)$ provides the least possible information about $Z$ relative to $g_{Z}$. When $g_{Z}(h(s))$ is degenerate (so $-H(g_{Z}(h(s)))=0$ is maximal), $h(s)$ provides the most possible information about $Z$ relative to $g_{Z}$.

Note that -- on this definition -- the amount of information $h(s)$ bears about $Z$ relative to a probe $g_{Z}$ is independent of whether the prediction $g_{Z}(h(s))$ is close to the true label distribution $p_{Z}(s)$ (indeed, it is possible that $p_{Z}(s)$ is uniform); we are interested only in the `uncertainty' of the probe's predictions.

It's worth comparing the notion of information employed here to that in the philosophical literature on representation. As \textcite{Shea2018-SHERIC} observes (p.78, footnote 5), (using my notation) his pointwise `correlational information'\footnote{More specifically, Shea's notion applies to `items' which can be in a range of `states' (pp.77-78). Formally, we can treat these items as random variables over the states they can be in, in order to enable comparison with the framework developed here.} between $\textbf{Z}$ and $\textbf{h}$ can be seen as a special case of the mutual information $I(\textbf{Z};\textbf{h})$, which doesn't pay attention to the underlying distribution over $\textbf{h}$. In order for this correlational information to be `exploitable', Shea requires both that the correlation holds across some regions of values for $\textbf{Z}$ and $\textbf{h}$, and that it holds for a `univocal' reason, where this latter condition is supposed to rule out merely coincidental correlations.

We can see that the existence of a successful probe $g_{Z}:h(D) \rightarrow \mathcal{P}(Z)$ demonstrates that both conditions hold. First, the `regions' in which the correlation holds are determined by the set of inputs $D$ comprising the task. Second, as Shea makes clear, the hallmark of a correlation that holds for a `univocal' reason is generalisation; there must be some pattern which allows prediction about the value of $\textbf{Z}$ using previously unseen states $h(s)$. By Definition \ref{successful probe}, a successful probe is exactly one that achieves below some threshold loss on an unseen test set, demonstrating generalisation.

So the successful probe $g_{Z}$ shows not only that there exists correlational information, but also that it is `exploitable', via a demonstration of one way in which the information could be decoded. When we talk about the degree of information relative to a particular probe across region $D$, then, we are effectively talking about the degree of correlational information that could be exploited \textit{using some particular decoder}. When we fix some particular activation vector $h(s)$, the probe's predictions $g_{Z}(h(s))$ are guesses for the value of $\textbf{Z}$, based on the correlation between $\textbf{Z}$ and $\textbf{h}$ that holds across region $D$; they are some particular first step towards exploiting this correlation on some particular input $s$, namely estimating $P_{\textbf{Z}}(\textbf{Z} \:|\: \textbf{h} = h(s))$. The degree of certainty in the probe's predictions (measured by the negative entropy) reflects the degree to which this correlational information could be exploited on \textit{this input},  using \textit{this decoder}.

What we care about is whether (and how) the downstream system actually exploits the information. Definition \ref{information relative to a probe} suggests the following step towards operationalising what it means for the system $S$ to use the information an activation vector $h(s)$ bears about property $Z$. The system counts as using the information $h(s)$ bears about $Z$ just when the following counterfactual holds: if we intervene on $h(s)$ to (i) reduce the information $h(s)$ bears about $Z$ \textit{relative to how the downstream system $S_{Dec}$ decodes the information $h(D)$ bears about $Z$} but (ii) leaving it otherwise unchanged, then the system's output $S(s)$ degrades (as measured by the task's goodness measure).

There are two challenges with further operationalising this idea, which I engage with in the following two subsections.

\subsubsection{How the Downstream System Decodes the Information}

The first challenge is to specify what it means to reduce the information $h(s)$ bears about $Z$ relative to how the downstream system $S_{Dec}$ decodes the information $h(D)$ bears about $Z$.

The problem here is that we don't know how the downstream system $S_{Dec}$ interacts with the information $h(D)$ bears about $Z$. In artificial neural systems, we know exactly how the downstream system interacts with the \textit{vehicles} of the representations, the activations themselves; that is, we can tell a complete low-level story about the function the system is computing. But this low-level story tells us very little about the way the system manipulates information about higher-level properties.

Note that things would be easier if we knew that the downstream system $S_{Dec}$ was a composition of a successful probe $g_{Z}$ and a function which took in the probe's predictions $g_{Z}(h(s))$ and performed some further computation. In this case, we could intervene on the informational content directly, simply by intervening on the probe's predictions. It's clear, though, that we can't think of the downstream system in this way; to the extent that it interacts with the information carried by the activations, it does so implicitly, manipulating information about many properties of inputs concurrently. In other words, we should not expect to find a single probe which uniquely identifies the way the downstream system decodes the information; it's not obvious how to apply Definition \ref{information relative to a probe} directly.

A solution to this challenge starts by recognising that it's essentially this difficulty that motivates the whole probing paradigm. If we could directly read off the predictions that neural models were making about high-level properties $Z$ from their activations, there would be little need for probing classifiers. Since we can't, though, we use probes as \textit{proxies} for the sorts of ways the downstream system could decode the information. This suggests a way of operationalising the idea of reducing the information $h(s)$ bears about $Z$ relative to the downstream system, namely to reduce the information relative to \textit{all} proxies (i.e. probes) for how the downstream system could decode the information.

Of course, we don't want just any probe $g_{Z}:h(D)\rightarrow \mathcal{P}(Z)$ to count as a proxy for how the downstream system could decode the information. First, we'll only be interested in \textit{successful} probes for the property, since we're interested in testing the counterfactual under the assumption that the downstream system is able successfully to manipulate information about the property. Second, we're only interested in probes which are \textit{plausible proxies} for how the downstream system decodes information about the property. I write the set of plausible proxies for how the downstream system $S_{Dec}$ decodes information about property $Z$ as $G_{Z, S_{Dec}}$. What does it mean for a probe $g_{Z}$ to be a member of $G_{Z, S_{Dec}}$?

Intuitively, it means that the probe $g_{Z}$ satisfies various constraints imposed by the downstream system $S_{Dec}$. Drawing on experimental practice, here are two plausible concrete constraints on $G_{Z, S_{Dec}}$:
\begin{definition}[Linearity Constraint]\hypertarget{linearity-constraint}{}
    If a probe $g_{Z} \in G_{Z, S_{Dec}}$, then $g_{Z}$ should be \textit{linear}. That is, we should be able to write
    \begin{equation*}
    g_{Z} = \texttt{softmax} \circ W_{g_{Z}}
    \end{equation*}
    where $W_{g_{Z}}$ is a linear transformation.
\end{definition}

\begin{definition}[Size Constraint]\hypertarget{size-constraint}{}
    If a probe $g_{Z} \in G_{Z, S_{Dec}}$, then there should be no probe $g'_{Z} \in G_{Z, S_{Dec}}$ such that the following two conditions hold:
    \begin{itemize}
        \item $g'_{Z}$ is at least as successful as $g_{Z}$, and
        \item $g'_{Z}$ is from the same family of models as $g_{Z}$, and is strictly `smaller' than $g_{Z}$.
    \end{itemize}
\end{definition}
\noindent `Smaller' in this second condition should be understood relative to the model family (i.e. rank of linear transformation for linear probes, hidden layer sizes for multi-layer perceptron probes (\cite{Shalev-Shwartz_Ben-David2014}), and so on).

The \hyperlink{linearity-constraint}{Linearity Constraint} is clearly motivated when the downstream system is linear, but even for highly non-linear downstream systems there is empirical evidence that decoding might be done linearly (\cite{elhage2021mathematical}; this claim is sometimes called the `Linear Representation Hypothesis'). The \hyperlink{linearity-constraint}{Size Constraint} is chosen to reflect experimental considerations on probe selection (see \cite{pimentel-etal-2020-pareto} for discussion on probe selection at the `Pareto frontier' of success and complexity).

The two constraints are not exhaustive,\footnote{Complexity notions from statistical learning theory could also be relevant here (e.g. bounding the Rademacher Complexity (\cite{Shalev-Shwartz_Ben-David2014}) of the class of functions compatible with the probe architecture with that compatible with $S_{Dec}$'s architecture).} and they should be understood as defeasible. A full characterisation of membership of $G_{Z, S_{Dec}}$ is beyond the scope of this paper. In general, I take it to be an open problem; membership of $G_{Z, S_{Dec}}$ will be refined by both theoretical and empirical, model-specific insights. I discuss this in Section \ref{probe selection}, where I highlight the role of probe selection and baselining in informing membership of $G_{Z, S_{Dec}}$. I take the identification of a distinctive role for `probes as proxies' (exemplified by $G_{Z, S_{Dec}}$) as contrasted with other roles probes play to be a conceptual contribution to the probing literature; I expand on this in Section \ref{probe selection}.

We are now ready to say what it means for an intervention to reduce the information $h(s)$ bears about $Z$ relative to how the downstream system $S_{Dec}$ decodes the information. I call this condition on interventions the \hyperlink{ablate condition}{\texttt{ablate} condition}.\footnote{\cite{lovering-pavlick-2022-unit} introduce an intervention with a similar spirit in a less general context. Although my operationalisation is very different from theirs, I adopt their terminology.} Note that since in my framework $h(s)\in \mathbb{R}^{d}$, an intervention on activation vectors is a function $a:\mathbb{R}^{d}\rightarrow \mathbb{R}^{d}$.
\begin{definition}[\texttt{ablate} condition]\hypertarget{ablate condition}{}
    An intervention $a:\mathbb{R}^{d}\rightarrow \mathbb{R}^{d}$ on activation vector $h(s)$ satisfies the \texttt{ablate} condition for property $Z$ iff
    for all successful $g_{Z} \in G_{Z, S_{Dec}}$:
    \begin{equation*}
        H(g_{Z}(a(h(s)))) > H(g_{Z}(h(s)))
    \end{equation*}
\end{definition}
\noindent So an intervention satisfies the \hyperlink{ablate condition}{\texttt{ablate} condition} if applying it to an activation vector reduces the information $h(s)$ bears about $Z$ relative to \textit{all successful proxies} $g_{Z} \in G_{Z, S_{Dec}}$ for the property (by the notion of information in Definition \ref{information relative to a probe}). The idea is that however the downstream system interacts with the information $h(s)$ bears about $Z$, an intervention satisfying the \hyperlink{ablate condition}{\texttt{ablate} condition} will reduce the information $h(s)$ bears about $Z$ relative to the downstream system.

\subsubsection{Leaving the Activation Vector Otherwise Unchanged}

The second challenge is to spell out what it means to leave the activation vector $h(s)$ `otherwise unchanged'. The problem is that the same vehicle $h(D)$ can bear correlational information about multiple properties $Z_{1},Z_{2},...$. So any intervention on $h(s)$ that modifies the information $h(s)$ bears about (e.g.) $Z_{1}$ could inadvertently modify the information it bears about (e.g.) $Z_{2}$. To the extent that we want representational claims about both $Z_{1}$ and $Z_{2}$ to feature in our explanations of the model, this is a problem.

To give a concrete example, suppose that BERT is performing a zero-shot subject/verb number agreement task). For the input $s$,
\begin{quote}
    \texttt{The man with the children [MASK] at us.}
\end{quote}
when predicting the masked token, BERT correctly assigns a higher probability to the singular verb form `waves' than to the plural verb form `wave'. We hypothesise that this is because BERT is representing the number of the nouns `man' and `children' (property $Z_{1}$), as well as a complete dependency parse of the sentence (property $Z_{2}$). Suppose we have evidence that $h(D)$ bears information about both $Z_{1}$ and $Z_{2}$ (e.g. by successfully training probes $g_{Z_{1}}$ and $g_{Z_{2}}$ to predict $Z_{1}$ and $Z_{2}$ respectively). To test our hypothesis, we apply an intervention to reduce the information $h(s)$ bears about number (i.e. the information about $Z_{1}$) relative to $g_{Z_{1}}$;\footnote{Of course, this intervention may fail to satisfy the \hyperlink{ablate condition}{\texttt{ablate} condition}, since it does not necessarily reduce the information relative to \textit{all} $g_{Z_{1}} \in G_{Z_{1},S_{Dec}}$. This doesn't matter for the purposes of this example.} we predict that BERT should now assign the singular and plural verb forms roughly equal probability, and this prediction comes true. Unbeknownst to us, though, we have modified $h(s)$ such that $g_{Z_{2}}$ now predicts the sentence as having a nonsensical dependency parse, undermining the claim that BERT uses the information $h(s)$ bears about $Z_{1}$ to perform its task. So the requirement that interventions be left otherwise unchanged is intended to protect the conclusions we can draw on the basis of interventions satisfying the \hyperlink{ablate condition}{\texttt{ablate} condition} from `false positives'.

The point is that \textit{distinct representational contents} can have the \textit{same representational vehicle}. It's important to emphasise that it is distinct from the related observation (made by \cite{feder-etal-2021-causalm}) that distinct linguistic properties $Z_{1}$ and $Z_{2}$ can be correlated (e.g. that $I(\textbf{Z$_{1}$}; \textbf{Z$_{2}$})>0$). To see this, note that it's perfectly possible for representational content about two correlated properties $Z_{1}$ and $Z_{2}$ to have different representational vehicles. It's also possible for the same pattern of activations $h(D)$ to bear information about two uncorrelated properties $Z_{1}$ and $Z_{2}$ (e.g. $Z_{1}$ is a semantic property and $Z_{2}$ an uncorrelated syntactic property).\footnote{\cite{feder-etal-2021-causalm} are concerned with a different (though related) problem to that of this paper. Rather than intervene on activations to evaluate representational claims, they tune an existing model so that its activations no longer bear information about a property (whilst preserving information about other properties).}

As above, we can use probes as proxies for ways the downstream system could decode information about properties (this time, information about properties we want to control for). This suggests a way of operationalising the idea of an intervention leaving information about properties `otherwise unchanged', namely the \hyperlink{control condition}{\texttt{control} condition}:
\begin{definition}[Control Condition]\hypertarget{control condition}{}
    An intervention $a:\mathbb{R}^{d}\rightarrow \mathbb{R}^{d}$ on activation vector $h(s)$ satisfies the \texttt{control} condition for property $Y$ iff for all successful $g_{Y} \in G_{Y, S_{Dec}}$:
    \begin{equation*}
        D_{KL}(g_{Y}(h(s)) \:||\: g_{Y}(a(h(s)))) \approx 0
    \end{equation*}
    where $D_{KL}$ denotes the Kullback–Leibler divergence (\cite{Cover_Thomas_2006}).
\end{definition}
\noindent In other words, an intervention on the information carried by $h(s)$ about property $Z$ relative to the downstream system $S_{Dec}$ satisfies the \hyperlink{control condition}{\texttt{control} condition} for property $Y\neq Z$ just in case it (approximately) preserves the predictions of probes for $Y$ (as measured by the Kullback–Leibler divergence, a familiar way of measuring changes in distributions). As stated, the \hyperlink{control condition}{\texttt{control} condition} is demanding; in practice, we would expect some relaxation of it. I discuss this in Section \ref{approximation}.

\subsubsection{Putting it Together: the \texttt{ablate} Intervention}

Having defined the \hyperlink{ablate condition}{\texttt{ablate}} and \hyperlink{control condition}{\texttt{control}} conditions, we are now ready to characterise the \hyperlink{ablate intervention}{\texttt{ablate} intervention}:
\begin{definition}[\texttt{ablate} Intervention]\hypertarget{ablate intervention}{}
    An intervention $a:\mathbb{R}^{d}\rightarrow \mathbb{R}^{d}$ on activation vector $h(s)$ is an \texttt{ablate} intervention for property $Z$ iff:
    \begin{itemize}
        \item $a$ satisfies the \hyperlink{ablate condition}{\texttt{ablate} condition} for property $Z$.
        \item $a$ satisfies the \hyperlink{control condition}{\texttt{control} condition} for all  properties $Y \neq Z$.
    \end{itemize}
\end{definition}
\noindent The \hyperlink{ablate intervention}{\texttt{ablate} intervention} operationalises the idea of reducing the information an activation vector bears about a property whilst leaving information about other properties unchanged, relative to the downstream system $S_{Dec}$. This suggests the following operationalisation of \hyperlink{use criterion}{Use}:
\begin{restatable}[Use Operationalised]{definition}{use-operationalised}\hypertarget{use-o}{}
The information $h(D)$ bears about $Z$ is used by system $S$
to perform task $D$ on input $s$ iff applying an \hyperlink{ablate intervention}{\texttt{ablate} intervention} to activation vector $h(s)$ degrades output $S(s)$ (as measured by the task's goodness measure on outputs) in an expected way.
\end{restatable}
\noindent It's worth saying something about what it means for an output to degrade `in an expected way'.\footnote{Thanks to an anonymous referee for suggesting this way of thinking about the issue.} The point is that we want a degradation to show that the information is being used \textit{to perform task $D$} (rather than having some other causal role in the system's behaviour). But degradation by itself is a weak requirement to meet; the output could degrade without indicating the sort of use we want to demonstrate. In general, in order for a degradation to count as `expected', it must plausibly relate to the property $Z$ we intervened on using the \hyperlink{ablate intervention}{\texttt{ablate} intervention}, and the way in which it degrades must reflect the intended effect of the \hyperlink{ablate intervention}{\texttt{ablate} intervention} (increasing uncertainty about the property). With respect to this second aspect, an anonymous referee gives the following example:
\begin{quote}
    Consider an autoregressive model that needs to predict what verb follows “the man with the children”. To conclude that the model encodes whether a verb’s subject is singular or plural, we should perform an ablation that results in the model doing worse at producing the right inflection for the verb (e.g. before ablation, it puts 99\% probability on “is” and 1\% on “are”, after ablation it puts 50\% on each). It would not be evidence of encoding singular/plural if the distribution instead changed to 99\% “was”, 1\% “were”, even if this distribution now resulted in degraded performance.
\end{quote}
The `in an expected way' clause is supposed to make sense of this intuition. Note that the sort of degradation expected is also dependent on the inputs $s$ themselves. Depending on the task $D$ and property $Z$, there will be many inputs $s \in D$ for which we know property $Z$ is not relevant for system performance (even if we think $Z$ is represented by the whole pattern of activations $h(D)$); on these inputs, we do not expect any degradation when we perform an \hyperlink{ablate intervention}{\texttt{ablate} intervention}. Part of establishing whether \hyperlink{use-o}{Use Operationalised} is satisfied is identifying which inputs $s \in D$ we should expect the system's performance to degrade on after we perform an \hyperlink{ablate intervention}{\texttt{ablate} intervention} on property $Z$.\footnote{As the referee points out, this makes sense of work on targeted evaluations, such as \textcite{wilcox-etal-2018-rnn}.}

What should we make of cases in which an output degrades, but in an unexpected way (for example, on an input $s$ for which we knew the property could not possibly be relevant)? I suggest that these cases provide evidence that the purported \hyperlink{ablate intervention}{\texttt{ablate} intervention} did not actually meet the \hyperlink{control condition}{\texttt{control} condition} (i.e. was not actually an \hyperlink{ablate intervention}{\texttt{ablate} intervention}); the intervention on property $Z$ has interfered with the information $h(s)$ bears about some other property $Y \neq Z$. Indeed, inputs like this provide a way of testing whether an intervention satisfies the \hyperlink{control condition}{\texttt{control} condition}.

The referee's example touches on another point, related to the degree of degradation. Ceteris paribus, the greater the degree of degradation (as measured by the task's goodness measure), the stronger the evidence that \hyperlink{use-o}{Use Operationalised} is satisfied; the point at which we allow that \hyperlink{use-o}{Use Operationalised} is satisfied will depend on our wider explanatory interests. Note that the `goodness' of the original output $S(s)$ is also relevant here; if we consider inputs on which the model failed to perform the task pre-intervention, then it's less clear how much degradation we should expect.\footnote{Thanks to another referee for discussion of this point.}

\subsection{Operationalising the Misrepresentation Criterion}\label{operationalising misrepresentation criterion}

In this subsection, I operationalise \hyperlink{misrepresentation criterion}{Misrepresentation}. As above, I do this via an intervention characterised by the predictions of probes.

Recall that \hyperlink{misrepresentation criterion}{Misrepresentation} was defined as follows:
\misrepresentation*
\noindent It's important to distinguish a misrepresentation of an input's property from the mere non-existence of a representation of the property. A pattern of activations $h(D)$ does not represent $Z$ when $h(D)$ bears no correlational information about $Z$ (i.e. when \hyperlink{information criterion}{Information} fails to be satisfied). By contrast, what happens in a case of misrepresentation is that $h(D)$ does bear correlational information about $Z$, but the activation vector $h(s)$ is interpreted by the downstream system as saying that $s$ has a label distribution distinct from the true label distribution $p_{Z}(s)$.

This suggests a step towards operationalising what it means for the activation vector $h(s)$ to misrepresent property $Z$. $h(s)$ counts as misrepresenting property $Z$ of input $s$ just when the following counterfactual holds: if we intervene on $h(s)$ to (i) make it such that the interpretation of the information $h(s)$ bears about $Z$ by the downstream system $S_{Dec}$ is closer to the true label distribution $p_{Z}(s)$ but (ii) leaving it otherwise unchanged, then the system's output $S(s)$ improves (as measured by the task's goodness measure). 

The second condition (leaving the activation vector `otherwise unchanged') should be familiar from the previous section: it is just the \hyperlink{control condition}{\texttt{control} condition}.

The first condition is harder to think about though. What does it actually mean for the interpretation of the information $h(s)$ bears about $Z$ by the downstream system to move closer to the true label distribution?

\subsubsection{Changing the Downstream System's Interpretation of the Label Distribution}

In discussing what it means to change the downstream system's `interpretation' of the information an activation vector bears about a property, we encounter the problem in the previous section, namely that the downstream system never explicitly interprets the information $h(s)$ bears about $Z$.

Before, I suggested resolving this using the predictions of probes for the property. Specifically, I suggested that we could see the information $h(s)$ bears about $Z$ being reduced relative to the downstream system if it was reduced relative to every \textit{successful proxy} for the downstream system. In this section, I employ the same basic insight, but with a different target (specifying an interpretation, rather than reducing information). The downstream system counts as interpreting an activation vector $h(s)$ as saying that $s$ has a certain property label distribution $q_{Z}(s) \in \mathcal{P}(Z)$ just in case every successful proxy for the downstream system predicts $q_{Z}(s)$ from $h(s)$. This suggests a way of understanding what it means for an intervention to modify the information $h(s)$ bears about $Z$ relative to the downstream system, namely the \hyperlink{modify condition}{\texttt{modify} condition}:
\begin{definition}[\texttt{Modify} condition]\hypertarget{modify condition}{}
    An intervention $a:\mathbb{R}^{d}\rightarrow \mathbb{R}^{d}$ on activation vector $h(s)$ satisfies the \texttt{modify} condition for property $Z$ and label distribution $q_{Z}(s) \in \mathcal{P}(Z)$ iff
    for all successful $g_{Z} \in G_{Z, S_{Dec}}$:
    \begin{equation*}
        D_{KL}(q_{Z}(s) \:||\: g_{Z}(a(h(s)))) < D_{KL}(q_{Z}(s) \:||\: g_{Z}(h(s)))
    \end{equation*}
    where $D_{KL}$ denotes the Kullback–Leibler divergence (\cite{Cover_Thomas_2006}).
\end{definition}
\noindent Intuitively, an intervention on activation vector $h(s)$ satisfies the \hyperlink{modify condition}{\texttt{modify} condition} for property $Z$ and distribution $q_{Z}(s)$ if it moves the predictions of successful proxies for the downstream system towards the distribution $q_{Z}(s)$, where `moves towards' is operationalised using the Kullback–Leibler divergence between the distribution $q_{Z}(s)$ and the proxy's predictions.

\subsubsection{The \texttt{modify} and \texttt{correct} Interventions}

Having defined the \hyperlink{modify condition}{\texttt{modify} condition}, we can characterise the \hyperlink{modify intervention}{\texttt{modify} intervention}:
\begin{definition}[\texttt{modify} Intervention]\hypertarget{modify intervention}{}
    An intervention $a:\mathbb{R}^{d}\rightarrow \mathbb{R}^{d}$ on activation vector $h(s)$ is a \texttt{modify} intervention for property $Z$ and label distribution $q_{Z}(s)$ iff:
    \begin{itemize}
        \item $a$ satisfies the \hyperlink{modify condition}{\texttt{modify} condition} for property $Z$ and label distribution $q_{Z}(s)$.
        \item $a$ satisfies the \hyperlink{control condition}{\texttt{control} condition} for all  properties $Y \neq Z$.
    \end{itemize}
\end{definition}
\noindent So the \hyperlink{modify intervention}{\texttt{modify} intervention} for property $Z$ and label distribution $q_{Z}(s)$ pushes the predictions of proxies for the downstream system towards $q_{Z}(s)$, whilst preserving the predictions of proxies on other properties.

Thinking of a representational explanation as an algorithm for manipulating high-level properties of inputs to derive the output $S(s)$ (something like Marr's algorithmic level of description, or the sense of explanation in \cite{Shea2018-SHERIC}; one way of making this more formal is to treat an explanation as a high-level causal model which `abstracts' the computation the system performs (\cite{geiger2019}, \cite{Geiger-Lu-2021})), we can use the \hyperlink{modify intervention}{\texttt{modify} intervention} to test explanations. Given a \hyperlink{modify intervention}{\texttt{modify} intervention} for property $Z$, explanations make different predictions about how outputs $S(s)$ (and the predictions of probes $g_{Y}$ for properties $Y \neq Z$ at later layers) should change, which can then be evaluated.\footnote{I intend this idea to be compatible with the causal abstraction framework (\cite{Geiger-Lu-2021}); indeed, it can be seen as a low-cost aid in the search for causal abstractions.}

The \hyperlink{correct intervention}{\texttt{correct} intervention} is an important species of \hyperlink{modify intervention}{\texttt{modify} intervention}:
\begin{definition}[\texttt{correct} Intervention]\hypertarget{correct intervention}{}
    An intervention $a:\mathbb{R}^{d}\rightarrow \mathbb{R}^{d}$ on activation vector $h(s)$ is a \texttt{correct} intervention for property $Z$ iff it is a \hyperlink{modify intervention}{\texttt{modify} intervention} for property $Z$ and the true label distribution $p_{Z}(s)$.
\end{definition}

\noindent The \hyperlink{correct intervention}{\texttt{correct} intervention} operationalises the idea of `correcting' the information that an activation vector bears about a property whilst leaving information about other properties unchanged (relative to the downstream system $S_{Dec}$). This suggests the following operationalisation of \hyperlink{misrepresentation criterion}{Misrepresentation}:

\begin{restatable}[Misrepresentation Operationalised]{definition}{misrepresentation-operationalised}\hypertarget{misrepresentation-o}{}
$h(s)$ misrepresents $Z$ iff applying a \hyperlink{correct intervention}{\texttt{correct} intervention} to activation vector $h(s)$ improves output $S(s)$ (as measured by the task's goodness measure on outputs) in an expected way.
\end{restatable}
\noindent Note that, as above, the `expected way' clause ensures that the behavioural effects are related to the performance of the task $D$; the improvement must plausibly relate to the property intervened on. In the previous subsection, I suggested focusing on inputs on which the system succeeds pre-intervention to evaluate \hyperlink{use-o}{Use Operationalised}; when evaluating \hyperlink{misrepresentation-o}{Misrepresentation Operationalised}, by contrast, we should focus on the system's failures, cases in which we would expect to see improvement post-intervention if the system were misrepresenting the property.\footnote{Again, thanks to an anonymous referee for discussion here.}

Consider \hyperlink{example 1}{Example 1} again. We were considering the claim that BERT mistakenly judges the following prompt $s$ as ungrammatical
\begin{quote}
    \texttt{John danced waltzes across the room.}
\end{quote}
because it \textit{misrepresents} `waltzes' as a verb. For each component $h$, we can test the explanation in terms of misrepresentation as follows.

Let $Z$ be a set of PoS tags. Here, the true label distribution $p_{Z}(s)$ is a degenerate distribution, which assigns probability $1$ to label $noun$ and probability $0$ to other labels. We apply a \hyperlink{correct intervention}{\texttt{correct} intervention} for property $Z$ to activation vector $h(s)$ and observe the effect on model behaviour. In order for the explanation in terms of $h(s)$ misrepresenting the PoS tag of `waltzes' to hold, BERT must assign a higher probability to the prompt being grammatical after performing the intervention.

\section{Discussion}\label{discussion}

Just as we can evaluate an operationalisation of a concept by how well it captures important features of the concept, we can also evaluate it by its empirical plausibility and usefulness. To this end, one central aim of this paper is to provide a common framework within which to compare and evaluate experimental evidence for making representational claims in contemporary NLP.

My operationalisation is idealised in many respects, which I discuss in Section \ref{approximation}. My hope is that it be thought of as an idealised experimental target, which different actual experiments will approximate to varying degrees; the degree of approximation serves as a metric for assessing an experiment's representational conclusions.\footnote{In other words, the idealisations have a \textit{normative} component (\cite{Colyvan2013-COLIIN}).} In Sections \ref{probe selection} and \ref{causal interventions}, I attempt to demonstrate this idea, showing that the framework I introduce can make sense of and assess several strands of empirical interpretability work, as well as suggesting directions for empirical research. Note that in this section I assume the system under discussion is an autoregressive transformer.

\subsection{Approximation}\label{approximation}

In this subsection, I outline some idealisations involved in my operationalisation, sketching how experiments could approximate them. Since this approximation comes in degrees (along several dimensions), this makes the support a particular experiment provides for a representational claim a matter of degree; each dimension provides an axis along which experimental evidence can be stronger or weaker. The threshold for when we want to call something a representation (that is, for when we judge the experimental evidence sufficient to justify the representational claim) will depend on our explanatory interests.

\subsubsection{How to Approximate `all' Successful Proxies?}

It's important that the \hyperlink{ablate condition}{\texttt{ablate}}, \hyperlink{control condition}{\texttt{control}} and \hyperlink{modify condition}{\texttt{modify}} conditions are defined relative to \textit{every way} the downstream system $S_{Dec}$ could plausibly interpret information about the properties. This is so we don't end up (in the case of the \hyperlink{ablate condition}{\texttt{ablate}} and \hyperlink{modify condition}{\texttt{modify}} conditions) with `false negatives', cases where the intervention fails to affect system behaviour because it doesn't affect the way the downstream system interprets the information about a property, or (in the case of the \hyperlink{control condition}{\texttt{control} condition}) `false positives', cases where the intervention affects system behaviour via affecting the system's interaction with information about an unrelated property. But how could an experiment ever verify that a condition is met for \textit{all} successful probes $g \in G_{Z,S_{Dec}}$?

Here's a general proposal for an approximation aimed at this dimension of idealisation. In general, given constraints imposed by $G_{Z,S_{Dec}}$, we should train multiple successful probes $g \in G_{Z,S_{Dec}}$ for each property (the property $Z$ we are intervening on and the properties $Y\neq Z$ we want to control for); the \hyperlink{ablate condition}{\texttt{ablate}}, \hyperlink{control condition}{\texttt{control}} and \hyperlink{modify condition}{\texttt{modify}} conditions should be satisfied relative to \textit{all} of these probes. We treat these probes as samples from the class of successful proxies for the downstream system; when we have taken sufficiently many samples, we count ourselves as having verified the condition holds (see my discussion of Iterative Null Space Projection (\cite{Ravfogel2020}) in Section \ref{iterated null space projection}). Note that not only is this standard practice in machine learning (c.f. cross-validation, where some finite number of different partitions of the task $D$ into training and test sets is used to draw an inference about all possible partitions; \cite{Bishop-2006}), training additional probes is feasible in practice, since additional probes do not require additional forward passes of the larger system $S$.

Another general technique for approximating this dimension of idealisation comes from \textcite{Tucker2022}, who propose a novel method for training probes, which works by zero-ing out a portion of randomly selected neurons from \textit{the input activation vector} (note: this is not standard dropout, in which \textit{intermediate} neurons are zero-ed out during training; \cite{Goodfellow-et-al-2016}) $h(s)$ during each training step of probes $g_{Z}$. The idea is that these `dropout probes' are more likely to detect all the ways in which information about a property $Z$ is encoded in $h(D)$, allowing them to serve as better approximations for the conditions I've introduced (in my framework, we can see this as suggesting that a single probe can approximate the behaviour of `all' $g_{Z} \in G_{Z, S_{Dec}}$).

\subsubsection{How to Approximate the \texttt{control} Condition?}

In order to satisfy the \hyperlink{control condition}{\texttt{control} condition}, an intervention must preserve the predictions of probes for \textit{all} properties $Y \neq Z$. There are two questions this raises:
\begin{itemize}
    \item First, how should we circumscribe the class of properties $Y \neq Z$ which we want to control for, given that we can't train probes for every property?
    \item Second, given that properties share a common representational vehicle $h(D)$, is it actually possible to change information about one property without affecting predictions about others? (This question is especially pressing when properties are correlated.)
\end{itemize}
Note the first question (how to identify possible `confounders'?) arises at some stage in the process for any attempt to evaluate counterfactuals about models by intervening on their activations. In all cases, then, we'll pick some finite subset of possible properties to control for. One benefit of the framework I've introduced is that, given a pattern of activations $h(D)$ and a property $Y$, it's cheap to probe for $Y$, meaning we can be liberal with the set of properties we control for (liberalism seems sensible here; there is no problem with controlling for redundant properties, whereas we don't want to miss relevant properties). In general, I take it that the set of properties we control for will depend on the sorts of representational explanations we want to evaluate (in other words, we will want to control for properties which feature in some plausible explanation of system behaviour); these explanations could come from domain-specific experts or -- as is increasingly common -- from the model itself, via appropriate prompting. Some properties we want to control for will doubtless be quite `alien', not reducible to existing concepts at a first glance.\footnote{Thanks to an anonymous referee for discussion on this point.} Various techniques exist for searching for these properties, via investigation of the geometry of the activation space (\cite{Chung_Abbott_2021}) or automated identification of shared features of sets of inputs which excite particular neurons (\cite{bills2023language}).

Having selected a finite set of properties to control for, there are also ways of approximating controlling for unconsidered properties, (e.g.) via appropriate regularisation (choosing interventions $a$ which satisfy the \hyperlink{ablate condition}{\texttt{ablate}} or \hyperlink{modify condition}{\texttt{modify}} conditions with the smallest possible distance between $a(h(s))$ and $h(s)$ on some metric).

To respond to the second question, it's true that we should expect any change to the vehicle $h(D)$ to affect the information $h(D)$ bears about properties $Y \neq Z$ (this is why the requirement in the \hyperlink{control condition}{\texttt{control} condition} is only that the predictions of probes $g_{Y}$ be \textit{approximately} unchanged); the degree of divergence is a dimension along which experimental evidence for claims about determinate representational content will come in degrees (presumably, we will care about controlling for some properties $Y$ more than others). That said, one advantage of my operationalisation is that the \hyperlink{ablate condition}{\texttt{ablate}} and \hyperlink{modify condition}{\texttt{modify}} conditions are relatively undemanding; they can be satisfied with small changes to the activation vector, implying that satisfying the \hyperlink{control condition}{\texttt{control} condition} might not be as difficult as it appears. Empirical work could investigate this.

When assessing how much violation of the \hyperlink{control condition}{\texttt{control} condition} to allow, one technique is to baseline the effect $a$ has on probe $g_{Y}$'s predictions against random interventions with the same property as $a$ (for example, which move $h(s)$ the same distance as measured by some metric). This idea makes sense of work by \textcite{Elazar2020}, who compare their proposed linear projection intervention to a baseline intervention into a random subspace with the same dimension as the subspace projected into by the original intervention. Indeed, comparing changes in predictions of successful probes for properties $Y_{1}$ and $Y_{2}$ in the face of random interventions on $h(s)$ is a way of testing how correlated the information that $h(D)$ bears about each property is. Another experimental technique in the NLP literature which could mitigate the problem of correlations between properties is to find a subset of inputs $s \in D$ in which the properties are de-confounded, and perform intervention experiments on this subset of the data (\cite{mccoy-etal-2019-right}); thanks to an anonymous referee for this suggestion.

Finally, to the extent that properties are correlated in the training data, there may be cases where we say that failure to be able to control for some property $Y \neq Z$ when intervening on the information about $Z$ (and vice-versa) shows that it is genuinely indeterminate whether $h(D)$ represents $Y$ or $Z$; this sort of indeterminacy is familiar to philosophers.

\subsubsection{How to Approximate Downstream System Behaviour?}

My operationalisation of \hyperlink{use criterion}{Use} and \hyperlink{misrepresentation criterion}{Misrepresentation} relies on performing an intervention and measuring changes to system outputs. This seems reasonable if the neurons in $h$ are close to the system's output layer, but one might worry if $h$ is taken from an early layer of the model that there is simply `too much going on' between $h(s)$ and $S(s)$ to interpret the result of an intervention (there could be back-up mechanisms in place which `wash out' the effect of the intervention, or the intervention could move the activation vector $h(s)$ too far away from its `usual' values $h(D)$ such that it generates strange system behaviour, an effect which compounds the more layers the downstream system has).

This worry generalises to any method which intervenes on neurons in an intermediate layer, including popular techniques like causal mediation analysis (\cite{vig2020_causalmediation}) or rank one model editing (\cite{meng2023locating}), but it's worth saying something about how it could be solved in practice.

One suggestion could be to employ a perspective from the empirical literature (the `logit lens'; see \cite{belrose2023eliciting} for a refinement on this, the `tuned lens') which treats each layer's output as though it was the system's pre-unembedding output, mapping it to a probability distribution over tokens using the model's own unembedding layer (\cite{Vaswani2017}); this perspective has empirical support, and sees the model as progressively refining its prediction at each layer. Rather than the whole system's behaviour, then, we could evaluate interventions by how they change the predictions of each downstream layer in turn. In effect, then, this proposal would treat each adjacent layer as though it were the downstream system (refining the set of proxies accordingly) and assessing \hyperlink{use-o}{Use} and  \hyperlink{misrepresentation-o}{Misrepresentation Operationalised} according to the intervention effect on the next layer's predictions. This could help us understand the role intermediate representations play in the system. Empirical work could investigate this.

\subsection{Probe Selection}\label{probe selection}

As observed above, there is a large literature on probe selection and baselining. This literature is motivated by the insight that -- since activations are (relatively) lossless encodings of inputs (\cite{Pimentel2020}) -- an overly expressive probe gives us little insight, since the probe may just `learn the task' (\cite{Hewitt-Liang-2019}). In my framework, we can identify two different roles for this literature.

\subsubsection{Probes as Proxies}

I have said relatively little about $G_{Z, S_{Dec}}$, the set of plausible proxies for how the downstream system could decode the information $h(D)$ bears about $Z$, apart from that it should be constrained by various properties of the downstream system $S_{Dec}$, such as its architecture. Probe selection can be seen exactly as examining what it takes to be a plausible proxy for $S_{Dec}$. A complete discussion of the sorts of constraints the probe selection literature places on $G_{Z, S_{Dec}}$ is left to future work, but I will give brief discussion of one aspect of probe selection which shows the value of framing probe selection in this way.

It is widely agreed that it is important for probes to \textit{generalise}, rather than \textit{memorise} (earlier, I made the same point in arguing that successful probes satisfy Shea's exploitability conditions). This is the reason why we train probes on a proper subset of $h(D)$, rather than the whole pattern of activations. Various techniques exist for encouraging and measuring generalisation (including standard regularisation techniques, such as dropout (\cite{srivastava14a}), restricting probe training data (\cite{zhang-bowman-2018}), and reporting probe `description length' (\cite{voita-titov-2020-information}; see their `online code') and `Bayesian mutual information' (\cite{pimentel-cotterell-2021-bayesian})). One popular baseline in this vein is `Control Tasks' (\cite{Hewitt-Liang-2019}), in which the gold labels $z \in Z$ for each input type in the probe's training and test data are randomly permuted, ensuring that a probe could only achieve success through memorisation (since it could not do better than random for unseen input types). Probes are selected only if no probe with the same architecture and hyperparameters is capable of memorisation (as measured by success on the control task).

In my framework, we can explain this work by seeing it as trying to refine $G_{Z, S_{Dec}}$. Specifically, the assumption is that the downstream system couldn't possibly be interacting with the information $h(D)$ bears about $Z$ through memorisation, implying the only probes we should treat as proxies are those which don't possibly memorise. By proposing techniques for measuring the degree of probe generalisation, this work is implicitly reducing the space of proxies for the downstream system.

\subsubsection{Probes as Locators of Information}

Much of this paper has been concerned with operationalising the notion of a pattern of activations $h(D)$ representing a property of an input. I haven't discussed, though, how one should select which component $h$ to perform experiments on.

One option is to perform experiments on every intermediate activation of the model; indeed, many experiments in the NLP literature do just this. Once experiments start involving interventions on activations, though, (especially those involving multiple properties) this becomes infeasible. How should we select parts of the model for more focused evaluation? In other words, which parts of the model are most appropriate to make representational claims about?

Once again, contemporary NLP practice is a guide here; we can use probes to help us choose candidate vehicles for representational claims. One will be able to train sufficiently expressive probes on many of the model's intermediate activations, but by more carefully selecting probes (specifically, by constraining probes' expressive power), one often ends up with a situation in which one can only train successful probes in some sequence of the model's intermediate layers.\footnote{For example, successful probes are trained at $k+1$ layers $l,l+1,..., l+k$, and probes trained at layers $<l$ or $>l+k$ are unsuccessful, for some $l,k$; exactly this picture can be found in (e.g.) \textcite{tenney2018what}.} This suggests -- as is recognised in NLP -- that the right part of the model to make representational claims about is in these intermediate layers, since information is more `easily extractable' at these layers (\cite{Pimentel2020}). Empirical work could test this intuition, using the framework I've introduced to operationalise representation.

Here we have a situation, then, where probe selection plays another role, namely to help locate parts of the model about which to make representational claims. It's important to emphasise that this is distinct from the role that probe selection plays in constraining proxies for the downstream system (i.e. in refining $G_{Z, S_{Dec}}$, on my picture). The two roles often go hand in hand, but are separable; for example, we might want to use a low-rank linear probe to draw attention to layers of the model in which information is most easily extractable, even though we believe that the downstream system could be interacting with the information non-linearly (i.e. that $G_{Z, S_{Dec}}$ could contain probes with non-linearities). I take it as a virtue of the framework I've introduced that it can distinguish between these roles.

\subsection{Causal Interventions on Model Activations}\label{causal interventions}

So far, I've said little about how we might find \hyperlink{ablate intervention}{\texttt{ablate}} or \hyperlink{modify intervention}{\texttt{modify}} interventions in practice. In this subsection, I outline examples of interventions on model activations from the NLP literature, showing that the idealised interventions I've introduced provide a useful framework for evaluating and comparing this extant work.

\subsubsection{Input-Based Methods}\label{input based methods}

The first techniques I discuss are input-based methods, such as causal mediation analysis (\cite{vig2020_causalmediation}; the most basic sort of `interchange intervention' also has this form (\cite{Geiger-Lu-2021})). The idea is to construct interventions on $h(s)$ for property $Z$ as follows: find inputs $s$ and $s'$ which differ with respect to property $Z$ (e.g. $p_{Z}(s)\neq p_{Z}(s')$) but which agree on properties $Y\neq Z$ (so $p_{Y}(s)=p_{Y}(s')$ for all $Y \neq Z$). We then define the intervention
\begin{equation*}
    a:h(s) \mapsto h(s')
\end{equation*}
So we substitute in the activation vector $h(s')$ (generated by input $s'$) for the activation vector $h(s)$ on input $s$, and see how this changes the system's behaviour. Using my framework, under the assumption that the two inputs vary only with respect to property $Z$, we appear to be guaranteed to have a \hyperlink{modify intervention}{\texttt{modify} intervention} for property $Z$ and distribution $p_{Z}(s')$.

Of course, input-based methods by themselves are limited to toy linguistic settings, where examples which differ only with respect to a single property are readily available (such as the gendered pronoun co-reference case in \cite{vig2020_causalmediation}). These settings constitute valuable domains for testing more general intervention methods (and in these settings, input-based methods may provide significant insight into how to define more general interventions, and where in the model to intervene), but it's less clear how to construct examples which differ only with respect to a single property in more general cases.

Even setting this limitation aside, there's a philosophical problem with using input-based methods to ground representational claims, which my framework can diagnose. In order for an input-based intervention to count as satisfying the \hyperlink{modify condition}{\texttt{modify} condition} for property $Z$ and distribution $p_{Z}(s')$, we have to already know that $h(s')$ \textit{does not misrepresent} property $Z$ of input $s'$. Similarly, in order for an input-based intervention to count as satisfying the \hyperlink{control condition}{\texttt{control} condition}, we have to accept that $h(s')$ represents properties $Y\neq Z$ in the same way as $h(s)$. This may be a plausible assumption (especially when the model gives the expected output $S(s')$), but nothing about the inputs themselves guarantees this.

Indeed, one of the main motivations for explaining neural models is precisely failures of robustness, often cases where inputs are indistinguishable to humans but receive different model outputs; one might think this is exactly the problem that an account of representation is supposed to solve. For example, how would an input-based method make sense of a case of misrepresentation, such as in \hyperlink{example 1}{Example 1}?
In a case of misrepresentation, (I've argued that) we want something like the following counterfactual to be true: `if the component had represented the property correctly, the system's output would have been better'. In the case of input-based methods, though, we get the antecedent `if the component had represented as it did in a different case', which isn't necessarily the same.

\subsubsection{Iterative Null Space Projection}\label{iterated null space projection}

The second technique I discuss is `Iterative Null Space Projection' (INLP; \cite{Ravfogel2020}, \cite{Elazar2020}, \textcite{lasri-etal-2022-probing}). INLP works by training a \textit{linear} probe
\begin{equation*}
    g_{Z} = \texttt{softmax} \circ W_{g_{Z}}
\end{equation*}
(where $W_{g_{Z}}$ is a linear transformation) to predict a property $Z$ from $h(D)$. For each input $s$, we intervene on $h(s)$ to project it into the kernel of the linear transformation $W_{g_{Z}}$. I write this projection intervention as $P_{Ker(W_{g_{Z}})}$. A new probe $g_{Z}^{(1)}$ is trained to predict $Z$ from $P_{Ker(W_{g_{Z}})}(h(D))$, and the process is repeated (this is the part of the technique which is `iterative') until convergence. The composition of these projection operations `$a$' is then applied to $h(s)$.

In the framework I've introduced, we can see INLP as approximating the \hyperlink{ablate intervention}{\texttt{ablate} intervention}, since it purports to reduce the information $h(s)$ bears about a property $Z$ for \textit{every} decoder of the information. Does it approximate the \hyperlink{ablate intervention}{\texttt{ablate} intervention} well enough to be used in \hyperlink{use-o}{Use Operationalised}? There are two points to make.

First, since INLP is only defined relative to linear probes, it implicitly sets $G_{Z, S_{Dec}}$ to include only linear probes (it adopts the \hyperlink{linearity-constraint}{Linearity Constraint}). To the extent that we think that some plausible proxies for the downstream system are non-linear, it will fail to satisfy even the \hyperlink{ablate condition}{\texttt{ablate} condition} (indeed, \textcite{Ravfogel2020} show that a non-linear probe can still be successfully trained to predict $p_{Z}(s)$ from $a(h(s))$).

Second, even assuming the the \hyperlink{linearity-constraint}{Linearity Constraint}, there's no reason to think that INLP satisfies the \hyperlink{control condition}{\texttt{control} condition}. To see this, consider \hyperlink{example 1}{Example 1} once again. Suppose that probe $g_{Z}$ decodes information about the PoS tag of `waltzes' and probe $g_{Y}$ decodes information about the PoS tag of `dances'. Even if $g_{Y}$ is also linear, applying $P_{Ker(g_{Z})}$ to activation vector $h(s)$ can affect the prediction of $g_{Y}$ on $h(s)$, namely when the range of the adjoint (the `rowspace') of the linear map $W_{g_{Z}}$ non-trivially intersects the range of the adjoint of $W_{g_{Y}}$ (\cite{axler}; this is of course guaranteed when the number of properties exceeds the activation vector dimension $d$). This is because there is a sense in which INLP is overly strong, which can be captured by my framework; it ensures not just that
\begin{equation*}
    H(g_{Z}(P_{Ker(g_{Z})}(h(s)))) > H(g_{Z}(h(s)))
\end{equation*}
as required by the \hyperlink{ablate condition}{\texttt{ablate} condition}, but that the resulting uncertainty is \textit{maximal}, since
\begin{equation*}
    g_{Z}(P_{Ker(g_{Z})}(h(s)))
\end{equation*}
will be uniform.

Note that even in those cases in which INLP fails to satisfy the \hyperlink{control condition}{\texttt{control} condition}, it might still be convenient to use INLP to approximate the \hyperlink{ablate intervention}{\texttt{ablate} intervention}, since it does give (coarse-grained) insight of the sort relevant for assessing \hyperlink{use-o}{Use Operationalised}. Furthermore, by better understanding the activation subspace in which interventions make a relevant difference to downstream system behaviour, we can design more fine-grained interventions (the `AlterRep' method in \cite{ravfogel-etal-2021-counterfactual} is an example of exactly this, designing an approximation of the \hyperlink{modify intervention}{\texttt{modify} intervention} for a binary property $Z$).

\subsubsection{Gradient-based Methods}\label{gradient-based methods}

The final technique I discuss intervenes on $h(s)$ using the gradient of the probe's prediction loss with respect to the input activation vector (\cite{Giulianelli2018}, \cite{Tucker2021}, \cite{Tucker2022}). Specifically, gradient-based methods compute
\begin{equation*}
    \nabla_{h(s)}L_{CE}(p_{Z}(s), g_{Z}(h(s)))
\end{equation*}
and update $h(s)$ using stochastic gradient descent (this process is repeated until some stopping condition obtains). So the update rule is as follows
\begin{equation*}
    h^{(k+1)}(s) \leftarrow h^{(k)}(s) - \alpha \nabla_{h(s)}L_{CE}(p_{Z}(s), g_{Z}(h(s)))
\end{equation*}
where $\alpha$ is the learning rate (a hyperparameter).

As used in the literature, $p_{Z}(s)$ is taken to be degenerate, and a gradient-based method is best thought of as an approximation of the \hyperlink{correct intervention}{\texttt{correct} intervention} for property $Z$ (albeit one using only a single probe $g_{Z} \in G_{Z, S_{Dec}}$). But it's worth observing that we could perform the gradient update with arbitrary target distributions, approximating arbitrary \hyperlink{modify intervention}{\texttt{modify} interventions}. Furthermore, there is no requirement the target distribution be degenerate (to my knowledge, this possibility has not been explored in the literature); by setting the target distribution as $Unif(Z)$ (the discrete uniform distribution over $Z$), we could satisfy the \hyperlink{ablate condition}{\texttt{ablate} condition}.

Do gradient-based interventions satisfy the \hyperlink{control condition}{\texttt{control} condition}? This depends on whether updating $h(s)$ affects the information $h(s)$ bears about properties $Y \neq Z$. This is largely an empirical question, but there is clearly nothing that prevents a gradient-based intervention from significantly affecting the information $h(s)$ bears about properties $Y \neq Z$.

Can we do better? If we look at the definition of the \hyperlink{control condition}{\texttt{control} condition} we note that what we want is to preserve the predictions of probes $g_{Y}$ for $Y \neq Z$, even as we intervene on $h(s)$ to change the predictions of $g_{Z}$. With a gradient-based method, there's a natural way to do this, namely \textit{build this extra condition directly into our loss term}. For each input $s$, I write the initial (pre-update) activation vector $h(s)$ as $h^{(0)}(s)$. Then an example of a loss term for target label $p(s)$ with respect to $g,g_{1},...,g_{n}$ after $k$ gradient-based updates to $h(s)$ could be:
\begin{align*}
    L(h^{(k)}(s)) &= L_{CE}(p_{Z}(s), g_{Z}(h^{(k)}(s))) +\\
    &\quad\sum_{Y \neq Z} \lambda_{Y} L_{CE}(g_{Y}(h^{(0)}(s)), g_{Y}(h^{(k)}(s)))
\end{align*}
where $\lambda_{Y}$ is a hyperparameter weighting the relative importance of property $Y$. After the first update to $h(s)$, these extra loss terms will work to penalise updates to $h(s)$ which shift the predictions the probes $g_{Y}$ for properties $Y \neq Z$. So we can see these extra loss terms as capturing exactly the way the \hyperlink{control condition}{\texttt{control} condition} was defined.

\section{Conclusion and Future Work}\label{conclusion}

This paper has operationalised a notion of representation suitable for contemporary NLP practice (and deep learning more generally).

By showing that a philosophically-informed notion of representation can be embedded into an NLP setting, I hope to have provided philosophers of cognitive science a gateway for applying well-developed theories of representation to contemporary debates in deep learning, as well as assisting them in using examples from contemporary NLP to compare theories of representation. As emphasised, the criteria I've operationalised do not by themselves constitute a full theory of representation (for example, they say little about how to fix determinacy of content). There is rich philosophical work to be done here, some of which was gestured at in Section \ref{three criteria}; my intention is that this paper gives philosophers a toolkit for translating fuller accounts of representation into a novel empirical setting, where they can help make better sense of machine learning practice.

Furthermore, as evidenced by my discussion of the grounding problem, philosophers are increasingly interested in questions in the development of AI, such as those related to understanding, explanation and agency. Not only do answers to many of these questions depend on claims about representation, but there is also consensus among philosophers that empirical work (especially interpretability work) should bear upon them; by organising this work and drawing out its philosophical implications, I hope that readers of the paper will be better-placed to engage with these issues. Finally, to the extent that `what the model represents' matters to downstream normative questions (as in \hyperlink{example 2}{Example 2}), this work should also be of interest to AI ethicists.

I anticipate that many of the more technical aspects of my proposal will be familiar to NLPers. Indeed, in Section \ref{discussion} I argued that something like this account is implicit in much contemporary interpretability work. My hope is that by presenting an explicit, unified framework for thinking about representation in NLP, I provide deep learning practitioners (and empirically-minded philosophers) with a useful idealisation against which to compare and assess a plethora of current interpretability techniques (especially in the probing literature), as evidenced by the discussion in Section \ref{discussion}. I take some concrete contributions to be:
\begin{itemize}
    \item outlining and operationalising a notion of \textit{misrepresentation}, which has received little explicit attention in the NLP literature;
    \item disambiguating between two sorts of effects interventions on activations can have with respect to how the downstream system interacts with information about a property;
    \item providing a common benchmark for comparing a large variety of distinct work for intervening on model activations;
    \item disambiguating between at least two roles for probe selection, and providing a way of understanding how work on probe selection informs more causal methods, via refinement of $G_{Z,S_{Dec}}$;
    \item suggesting (to my knowledge) a novel method for controlling for confounding properties, namely by preserving the predictions of probes for these confounders, and showing that this can be accommodated by a gradient-based approach.
\end{itemize}
More generally, I hope that this paper demonstrates the role that philosophers of science can play in contributing to debates in the development of AI, many of which -- as the use of AI proliferates -- have implications for important normative questions in society at large.

\newpage

\section*{Acknowledgements}

Thanks especially to Rosa Cao and Thomas Icard for detailed, insightful comments at every stage of this project. Part of this work was presented at the NYU/Columbia Philosophy of Deep Learning Conference; thanks to participants there for helpful discussion, in particular Nicholas Shea, Raphaël Millière, Cameron Buckner, Anders Søgaard, Patrick Butlin and Tal Linzen. Thanks also to Cameron Domenico Kirk-Giannini, J. Dmitri Gallow, Simon Goldstein, N.G. Laskowski, Nathaniel Sharadin, Ben Levinstein, William D’Allesandro, Robert Long, Frank Hong, Elliot Thornley and Mitchell Barrington for feedback on an earlier draft. Finally, thanks to two anonymous referees for the BJPS for a range of thoughtful suggestions. This work was supported by Stanford’s HAI graduate fellowship and completed during the CAIS Philosophy Fellowship.

\printbibliography

\end{document}